\documentclass[10pt,twocolumn,letterpaper]{article}

\usepackage{graphics}  
\usepackage{epstopdf}
\usepackage{caption}
\usepackage{subfigure}
\usepackage{multirow}
\usepackage{array}
\usepackage{cvpr}
\usepackage{times}
\usepackage{epsfig}
\usepackage{graphicx}
\usepackage{amsmath}
\usepackage{amssymb}
\usepackage{amsthm,amsmath,amssymb}
\usepackage{mathrsfs}
\usepackage{booktabs}
\usepackage{authblk}

\usepackage{enumitem}
\setitemize[1]{itemsep=-2pt,partopsep=-2pt,parsep=\parskip,topsep=-2pt}

\usepackage[pagebackref=true,breaklinks=true,letterpaper=true,colorlinks,bookmarks=false]{hyperref}

\cvprfinalcopy 

\def\cvprPaperID{3225} 

\ifcvprfinal\fi
\begin{document}

\title{General Instance Distillation for Object Detection}
\makeatletter
\renewcommand\AB@affilsepx{\quad \protect\Affilfont}
\renewcommand*{\@fnsymbol}[1]{\ensuremath{\ifcase#1\or *\or \dagger\or \ddagger\or
    \mathsection\or \mathparagraph\or \|\or **\or \dagger\dagger
    \or \ddagger\ddagger \else\@ctrerr\fi}}
\makeatother
\newcommand*\samethanks[1][\value{footnote}]{\footnotemark[#1]}

\author[1]{Xing Dai\thanks{The first two authors contribute equally and the order is alphabetical.}}
\author[1,2]{Zeren Jiang\samethanks\thanks{This work was done when Zeren was an intern at MEGVII Tech.}}
\author[1]{Zhao Wu}
\author[1]{\\Yiping Bao}
\author[1]{Zhicheng Wang}
\author[2]{Si Liu}
\author[1]{Erjin Zhou}

\affil[1]{MEGVII Technology}
\affil[2]{BeiHang University}

\affil[ ]{\protect\\
\tt\small daixinghome@gmail.com \quad
\tt\small zeren.jiang99@gmail.com \quad
\tt\small wuzhao@megvii.com \quad \protect\\
\tt\small baoyiping@megvii.com \quad
\tt\small wangzhicheng@megvii.com \quad
\tt\small liusi@buaa.edu.cn \quad
\tt\small zej@megvii.com
}

\setlength{\affilsep}{0em}
\renewcommand\Authsep{\quad}
\renewcommand\Authand{\quad}
\renewcommand\Authands{\quad}

\maketitle

\begin{abstract}
   In recent years, knowledge distillation has been proved to be an effective solution for model compression. This approach can make lightweight student models acquire the knowledge extracted from cumbersome teacher models. However, previous distillation methods of detection have weak generalization for different detection frameworks and rely heavily on ground truth (GT), ignoring the valuable relation information between instances. Thus, we propose a novel distillation method for detection tasks based on discriminative instances without considering the positive or negative distinguished by GT, which is called general instance distillation (GID). Our approach contains a general instance selection module (GISM) to make full use of feature-based, relation-based and response-based knowledge for distillation. Extensive results demonstrate that the student model achieves significant AP improvement and even outperforms the teacher in various detection frameworks. Specifically, RetinaNet with ResNet-50 achieves 39.1\% in mAP with GID on COCO dataset, which surpasses the baseline 36.2\% by 2.9\%, and even better than the ResNet-101 based teacher model with 38.1\% AP.
\end{abstract}

\section{Introduction}

In recent years, the accuracy of object detection has made a great progress due to the blossom of deep convolutional neural network (CNN). The deep learning network structure, including a variety of one-stage detection models \cite{liu2016ssd,redmon2016you,Redmon_2017_CVPR,redmon2018yolov3,Lin_2017_ICCV} and two-stage detection models \cite{NIPS2015_5638,Lin_2017_CVPR,he2017mask,cai2018cascade}, has replaced the traditional object detection and has become the mainstream method in this field. Furthermore, the anchor-free frameworks \cite{law2018cornernet,duan2019centernet,Tian_2019_ICCV} have also achieved better performance with more simplified approaches. However, these high-precision deep learning based models are usually cumbersome, while a lightweight with high performance model is demanded in practical applications. Therefore, how to find a better trade-off between the accuracy and efficiency has become a crucial problem.

\begin{figure} 
  \centering
  \includegraphics[width = \linewidth]{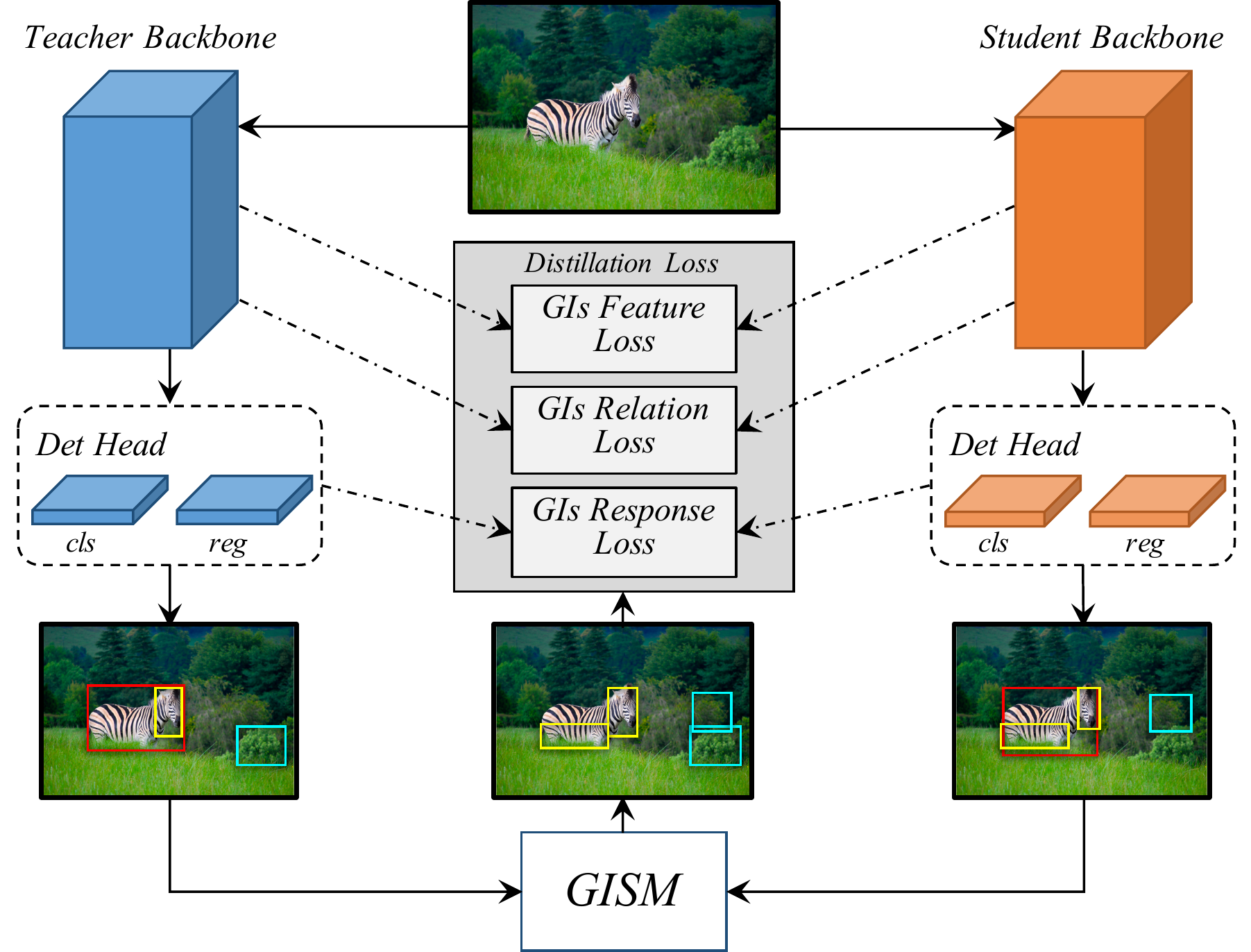} 
  \caption{Overall pipeline of general instance distillation (GID). General instances (GIs) are adaptively selected by the output both from teacher and student model. Then the feature-based, relation-based and response-based knowledge are extracted for distillation based on the selected GIs.}
  \label{GID_framework}
\end{figure}

Knowledge Distillation (KD), proposed by Hinton et al. \cite{hinton2015distilling}, is a promising solution for the above problem. Knowledge distillation is to transfer the knowledge of large model to small model, thereby improving the performance of the small model and achieving the purpose of model compression. At present, the typical forms of knowledge can be divided into three categories \cite{gou2020knowledge}, response-based knowledge \cite{hinton2015distilling,Park_2019_CVPR}, feature-based knowledge \cite{romero2014fitnets,Yim_2017_CVPR,DBLP:journals/corr/abs-1811-03233} and relation-based knowledge \cite{Park_2019_CVPR,Liu_2019_CVPR,tian2019contrastive,Tung_2019_ICCV, li2020local}. However, most of the distillation methods are mainly designed for multi-class classification problems. Directly migrating the classification specific distillation method to the detection model is less effective, because of the extremely unbalanced ratio of positive and negative instances in the detection task. Some distillation frameworks designed for detection tasks cope with this problem and achieve impressive results, \eg Li et al. \cite{Li_2017_CVPR} address the problem by distilling the positive and negative instances in a certain proportion sampled by RPN, and Wang et al. \cite{Wang_2019_CVPR} further propose to only distill the near ground truth area. Nevertheless, the ratio between positive and negative instances for distillation needs to be meticulously designed, and distilling only GT-related area may ignore the potential informative area in the background. Moreover, current detection distillation methods cannot work well in multi detection frameworks simultaneously, \eg  two-stage, anchor-free methods. Therefore, we hope to design a general distillation method for various detection frameworks to use as much knowledge as possible effectively without concerning the positive or negative.

Towards this goal, we propose a distillation method based on discriminative instances, utilizing response-based knowledge, feature-based knowledge as well as relation-based knowledge, as shown in Fig \ref{GID_framework}. There are several advantages:
(i) \begin{itshape}We can model the relational knowledge between instances in one image for distillation.\end{itshape}
Hu et al. \cite{Hu_2018_CVPR} demonstrates the effectiveness of relational information on detection tasks. However, the relation-based knowledge distillation in object detection has not been explored yet.
(ii) \begin{itshape}We avoid manually setting the proportion of the positive and negative areas or selecting only the GT-related areas for distillation.\end{itshape}
Though GT-related areas are almost informative, the extremely hard and simple instances may be useless, and even some informative patches from the background can be useful for students to learn the generalization of teachers. Besides, we find that the automatic selection of some discriminative instances between the student and teacher for distillation can make knowledge transferring more effective. Those discriminative instances are called general instances (GIs), since our method does not care about the proportion between positive and negative instances, nor does it rely on GT labels.
(iii) \begin{itshape}Our methods have robust generalization for various detection frameworks.\end{itshape}
GIs are calculated upon the output from student and teacher model without relying on certain modules from a specific detector or some key characteristic, such as anchor, from a particular detection framework.

To sum up, this paper makes the following contributions:
\begin{itemize}
  \item Define general instance (GI) as the distillation target, which can effectively improve the distillation effect of the detection model.
  \item 
  Based on GI, we first introduce the relation-based knowledge for distillation on detection tasks and integrate it with response-based and feature-based knowledge, which makes student surpass the teacher.
  \item 
  We verify the effectiveness of our method on the MSCOCO \cite{10.1007/978-3-319-10602-1_48} and PASCAL VOC \cite{Everingham10thepascal} datasets, including one-stage, two-stage and anchor-free methods, achieving state-of-the-art performance.
\end{itemize}
\section{Related Work}

\subsection{Object Detection}

The current mainstream object detection algorithms are roughly divided into two-stage and one-stage detectors. Two-stage methods \cite{Lin_2017_CVPR,he2017mask,cai2018cascade} represented by Faster R-CNN \cite{NIPS2015_5638} maintain the highest accuracy in the detection field. These methods utilize region proposal network (RPN) and refinement procedure of classification and location to obtain better performance. However, high demands for lower latency bring one-stage detectors \cite{liu2016ssd,redmon2016you} under the spotlight, which achieve classification and location of targets through the feature map directly.

In recent years, another criterion divides detection algorithm into anchor-based and anchor-free methods. Anchor-based detectors such as \cite{Redmon_2017_CVPR,Lin_2017_ICCV,liu2016ssd} solve object detection tasks with the help of anchor boxes, which can be viewed as pre-defined sliding windows or proposals. Nevertheless, all anchor-based methods need to be meticulously designed and calculate a large number of anchor boxes which takes much computation. To avoid tunning hyper-parameters and calculation related to anchor boxes, anchor-free methods  \cite{redmon2016you,law2018cornernet,duan2019centernet,Tian_2019_ICCV} predict several key points of target, such as center and distance to boundaries, reach a better performance with less cost. 

\subsection{Knowledge Distillation}

Knowledge distillation is a kind of model compression and acceleration approach which can effectively improve the performance of small models with guiding of teacher models. In knowledge distillation, knowledge takes many forms, \eg the soft targets of the output layer \cite{hinton2015distilling}, the intermediate feature map \cite{romero2014fitnets}, the distribution of the intermediate feature \cite{huang2019like}, the activation status of each neuron \cite{DBLP:journals/corr/abs-1811-03233}, the mutual information of intermediate feature \cite{Ahn_2019_CVPR}, the transformation of the intermediate feature \cite{Yim_2017_CVPR} and the instance relationship \cite{Park_2019_CVPR,Liu_2019_CVPR,tian2019contrastive,Tung_2019_ICCV}. Those knowledge for distillation can be classified into the following categories \cite{gou2020knowledge}: response-based \cite{hinton2015distilling}, feature-based \cite{romero2014fitnets,huang2019like,DBLP:journals/corr/abs-1811-03233,Ahn_2019_CVPR,Yim_2017_CVPR}, and relation-based \cite{Park_2019_CVPR,Liu_2019_CVPR,tian2019contrastive,Tung_2019_ICCV}.

Recently, there are some works applying knowledge distillation to object detection tasks. Unlike the classification tasks, the distillation losses in detection tasks will encounter the extreme unbalance between positive and negative instances. Chen et al. \cite{NIPS2017_6676} first deals with this problem by underweighting the background distillation loss in the classification head while remaining imitating the full feature map in the backbone. Li et al. \cite{Li_2017_CVPR} designs a distillation framework for two-stage detectors, applying the L2 distillation loss to the features sampled by RPN of student model, which consists of randomly sampled negative and positive proposals discriminated by ground truth (GT) labels in a certain proportion. Wang et al. \cite{Wang_2019_CVPR} proposes a fine-grained feature imitation for anchor-based detectors, distilling the near objects regions which are calculated by the intersection between GT boxes and anchors generated from detectors. That is to say, the background areas will hardly be distilled even if it may contain several information-rich areas. Similar to Wang et al. \cite{Wang_2019_CVPR}, Sun et al. \cite{sun2020distilling} only distilling the GT-related region both on feature map and detector head.

In summary, the previous distillation framework for detection tasks all manually set the ratio between distilled positive and negative instances distinguished by the GT labels to cope with the disproportion of foreground and background area in detection tasks. Thus, the main difference between our method and the previous works can be summarized as follows: (i) Our method does not rely on GT labels, nor does it care about the proportion between positive and negative instances selected for distillation. It is the information gap between student and teacher that guides the model to choose the discriminative patches for imitation. (ii) None of the previous methods take advantage of the relation-based knowledge for distillation. However, it is widely acknowledged that the relation between objects contains tremendous information even within one single image. Thus, based on our selected discriminative patches, we extract the relation-based knowledge among them for distillation, achieving further performance gain.

\section{General Instance Distillation}

Previous work \cite{Wang_2019_CVPR} proposed that the feature regions near objects have considerable information which is useful for knowledge distillation. However, we find that not only the feature regions near objects but also the discriminative patches even from the background area have meaningful knowledge. Base on this finding, we design the general instance selection module (GISM), as shown in Fig \ref{GISM_fig}. The module utilizes the predictions from both teacher and student model to select the key instances for distillation. 

\begin{figure*} 
    \centering
    \subfigure[Feature-based and relation-based distillation]{
    \includegraphics[width=0.48\textwidth]{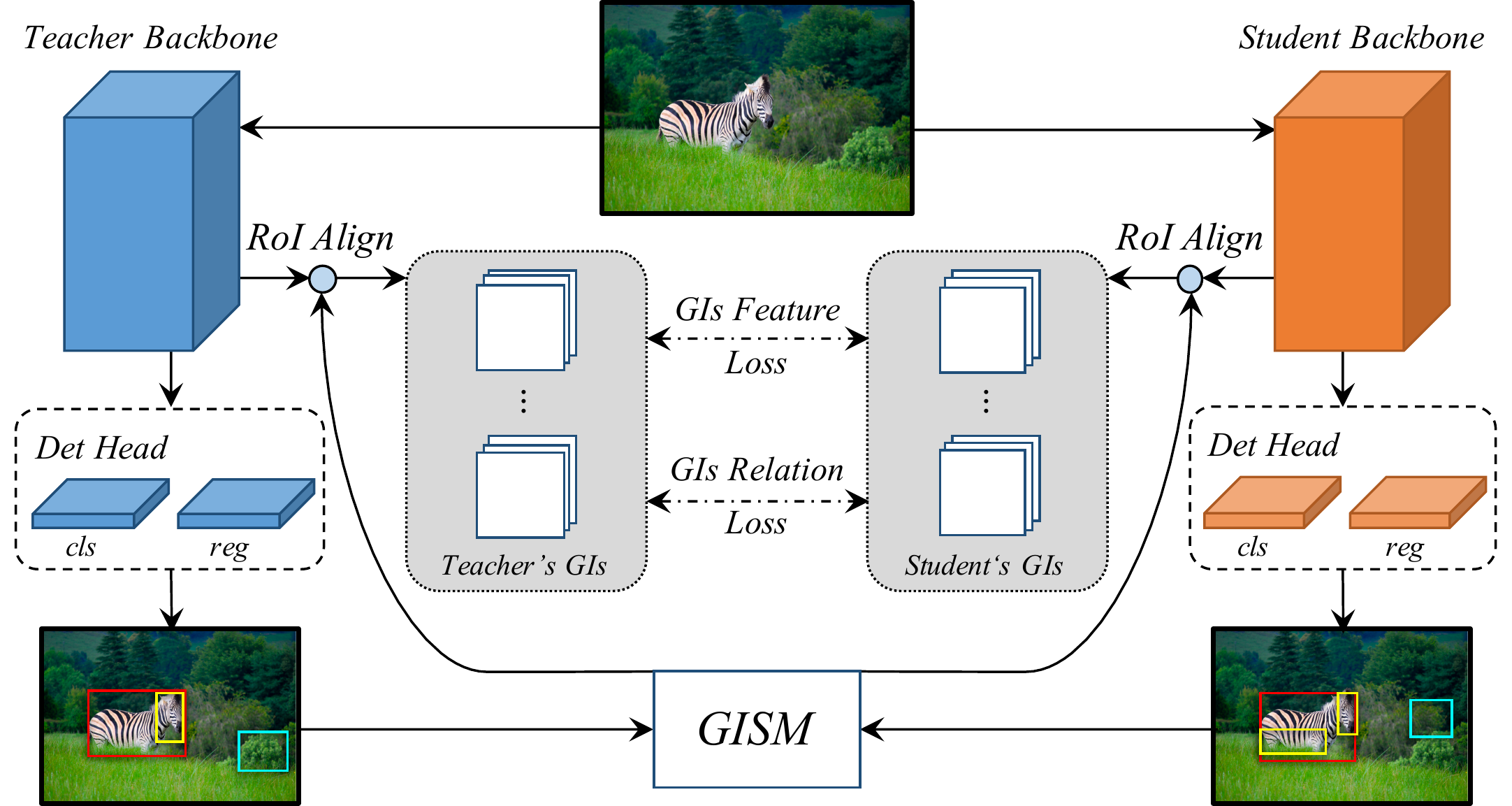}
        \label{Feature-Relation_distillation}
    }
    \subfigure[Response-based distillation]{
	    \includegraphics[width=0.48\textwidth]{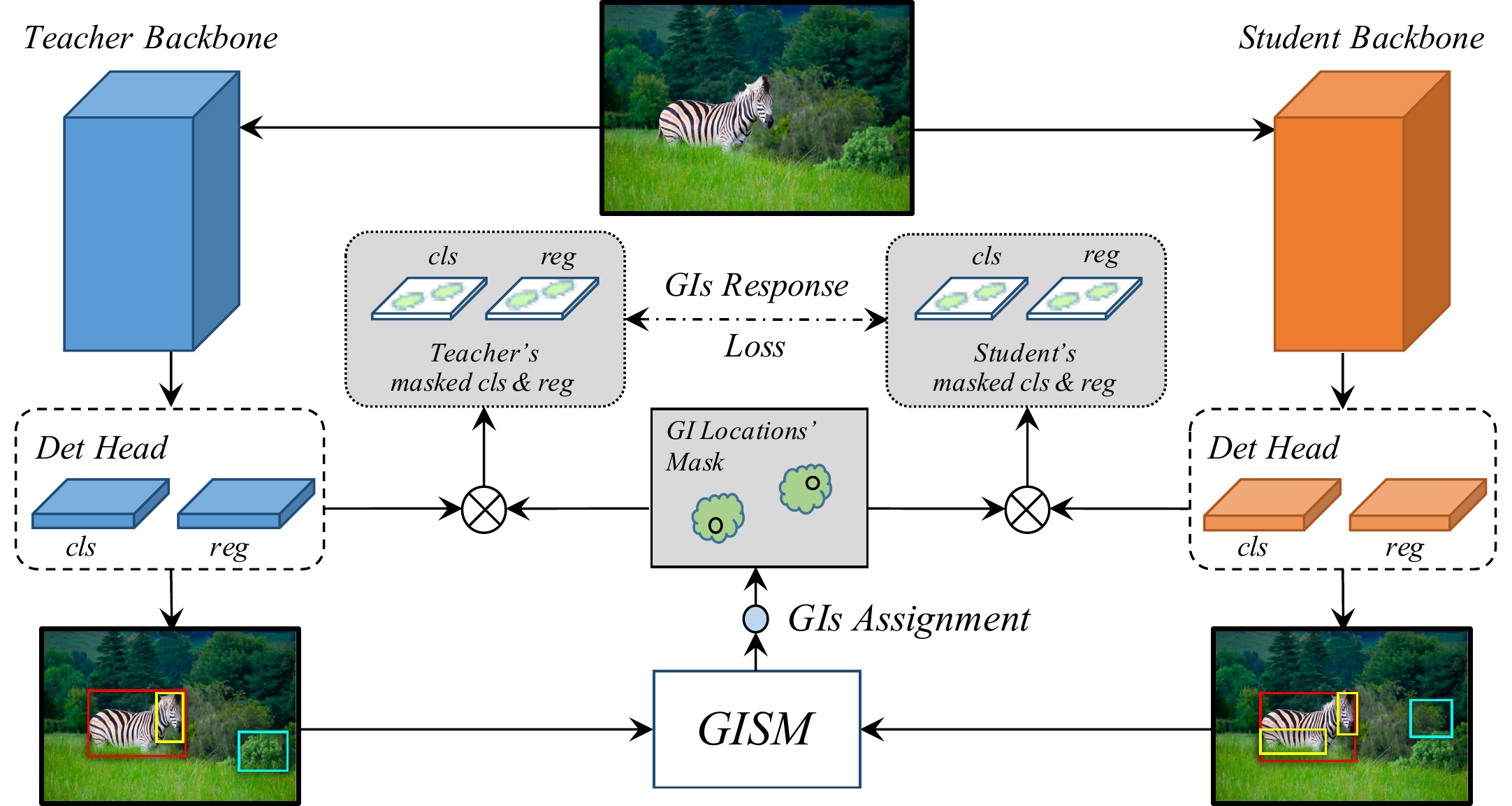}
        \label{Response_distillation}
    }
    \caption{Details of our method: (a) Selected GIs are used to crop the feature in student and teacher backbone by ROI Align. Then the feature-based and relation-based knowledge are extracted for distillation. (b) Selected GIs first generate a mask by GIs assignment. Then masked classification and regression head are distilled to utilize response-based knowledge.}
    \label{main_fig}
\end{figure*}

Furthermore, to make better use of the information provided by the teacher, we extract and take advantage of feature-based, relation-based and the response-based knowledge for distillation, as shown in Fig \ref{main_fig}. The experimental results show that our distillation framework is general for current state-of-the-art detection models.

\subsection{General Instance Selection Module}

\begin{figure} 
    \centering
    \includegraphics[width = \linewidth]{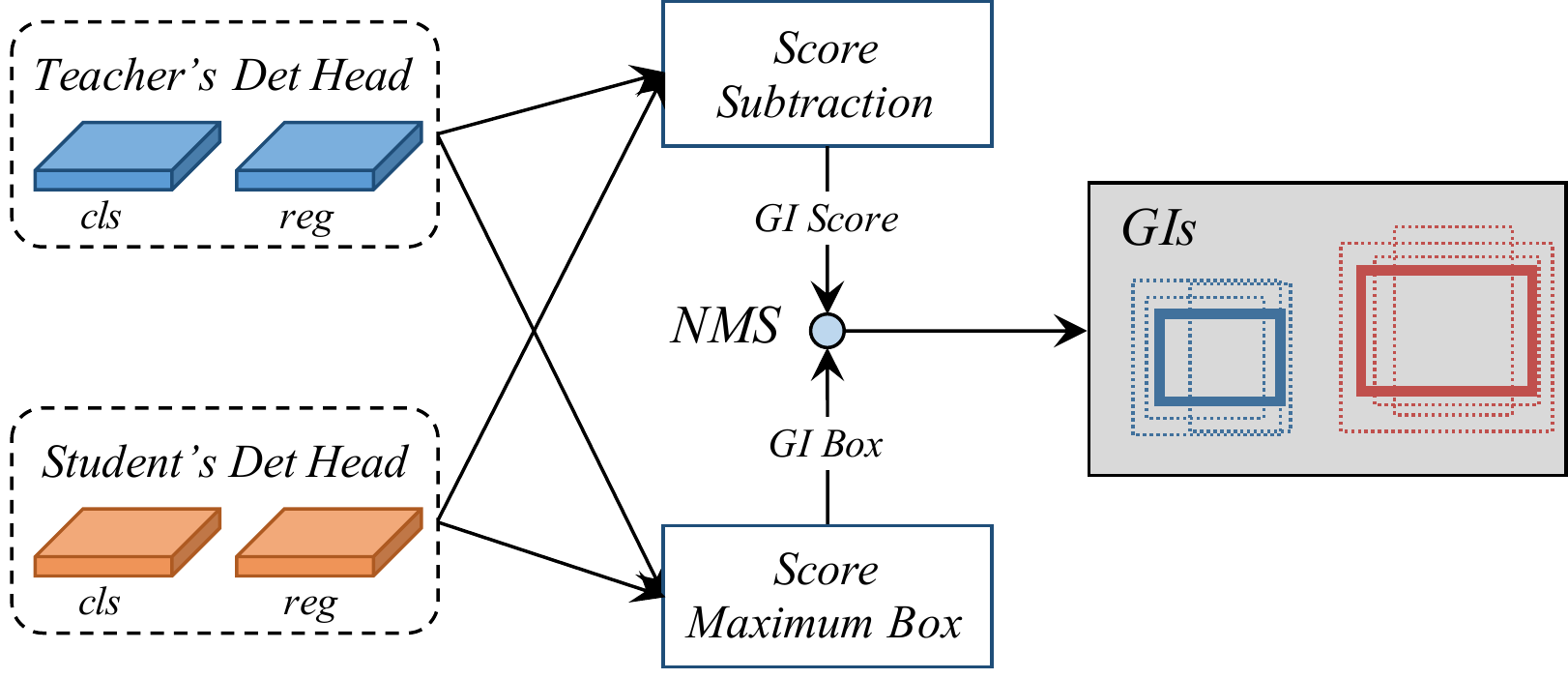} 
    \caption{Illustration of the general instance selection module (GISM). To obtain the most informative locations, we calculate the L1 distance of classification scores from student and teacher as GI scores, and preserve regression boxes with higher scores as GI boxes. To avoid losses double counting, we use the non-maximum suppression (NMS) algorithm to remove duplicates.}
    \label{GISM_fig}
\end{figure}

In detection model, predictions indicate the attention patches which are commonly meaningful areas. The difference of such patches between teacher and student model is also closely related to their performance gap. In order to quantify the difference for each instance and then select the discriminative instances for distillation, we propose two indicator: GI score and GI box. Both of them are dynamically calculated during each training step. For saving the computation resources during training, we simply calculate the L1 distance of classification score as GI score and choose box with higher score as GI box. Fig \ref{GISM_fig} illustrates the procedure of generating GI, and the score and box of which from each predicted instance $r$ is defined as below.
{\setlength\abovedisplayskip{6pt}
\setlength\belowdisplayskip{0pt}
\begin{align}
    P_{GI}^{r} &= \max\limits_{0<c{\leq}C}  \left| P_{t}^{rc}  - P_{s}^{rc}  \right|, \\
    B_{GI}^{r} &= \left\{
        \begin{array}{lll}
            B_{t}^{r},      & \max\limits_{0<c{\leq}C}{P_{t}^{rc} > \max\limits_{0<c{\leq}C}P_{s}^{rc}} \\
            B_{s}^{r},      & \max\limits_{0<c{\leq}C}{P_{t}^{rc} \leq \max\limits_{0<c{\leq}C}P_{s}^{rc}} \\
        \end{array} \right. ,\\
    {GI} &= NMS(P_{GI}, B_{GI}),
\end{align}
}
\par\noindent where $P_{GI}$ and $B_{GI}$ denote GI score and GI box. 
For one-stage detectors, $P_{t}$ and $P_{s}$ are the classification score predicted by the teacher and student separately. As for two-stage detectors, $P$ refers to the objectness score predicted by RPN. Meanwhile, $B_{t}$ and $B_{s}$ are the regression boxes predicted by the teacher and student, corresponding to the score $P_{t}$ and $P_{s}$. $R$ is the number of the predicted boxes and $C$ is the number of classes. $r$, $c$ are indexes in the dimension of $R$, $C$ in subject to $0{<}c{\leq}C$ and $0{<}r{\leq}R$. Since we set the detection heads of teacher and student model pair to be exactly the same, so these two networks have equal $R$ number of prediction boxes with one-to-one corresponding location.

Though we identified the indicator of GI scores and corresponding boxes, these instances with high GI scores  are likely to be highly overlapped, thus leading to distillation loss double counting. To deal with these redundant and correlated regions, we use standard non-maximum suppression (NMS) to perform deduplication. Given a list of instances with GI scores and boxes, NMS works by iteratively selecting the instance with the highest GI score, and then removing all lower GI score instances that have high overlap with the selected region. We use an IoU threshold of 0.3 to select dispersing instances. Moreover, only top $K$ instances with the highest score are chosen as the final GI for distillation in each image.

\subsection{Feature-based Distillation}
Most of the SOTA detection models have introduced the Feature Pyramid Networks (FPN) \cite{Lin_2017_CVPR}, which can significantly improve the robustness of multi-scale detection. Since the FPN combines the feature of multiple backbone layers, we intuitively choose the FPN for distillation. To be specific, we crop the feature from the matching FPN layer according to the different size of each GI box.

Given that the target sizes vary greatly in detection tasks, directly performing pixel-wise distillation will make the model more incline to learn large targets. Therefore, as shown in Fig \ref{Feature-Relation_distillation}, we adopt the ROIAlign \cite{he2017mask} algorithm, which resizes GI feature of different sizes to the same size and then perform distillation, treating each target equally. The feature-based distillation loss is as follows:
{\setlength\abovedisplayskip{6pt}
\setlength\belowdisplayskip{0pt}
\begin{align}
    L_{Feature}&=\frac{1}{K}\sum_{i=1}^{K}{\left \| t_{i}-s^\prime_{i}\right \|_{2}^{2}},
    \\
    s^\prime &= f_{adapt}(s),
\end{align}
}
\par\noindent in which $K$ is the number of GI selected by GISM with top $K$ GI scores, $t_i$ and $s_i$ are the $i^{th}$ GI feature extracted from the teacher and student model by ROIAlign algorithm, $f_{adapt}$ is the linear adaptation function to adapt $s_i$ to the same dimension as $t_i$.

\subsection{Relation-based Distillation}

Relational information \cite{Park_2019_CVPR,Liu_2019_CVPR} between different objects has played a significant role in distillation for classification task. However, the relation-based knowledge distillation for detection tasks remains unexplored. Since the instances in the same scene are highly correlated, regardless of foreground or background, this correlation information can help the student network converge more effectively.

Owe to the informative GIs selected by GISM, we are able to take full advantage of the correlation between discriminative instances. Only performing one-to-one feature distillation is certainly not enough to import more knowledge. Therefore, to mine the valuable relation knowledge underlying a batch of GIs, we further introduce relation-based knowledge for distillation. Here we use Euclidean distance to measure the relevance of 
instances, and L1 distance to transfer knowledge. As shown in Fig \ref{Feature-Relation_distillation}, we additionally utilize the correlation information between GIs to distill knowledge from teacher to student. The loss expression is as follows:

{\setlength\abovedisplayskip{0pt}
\setlength\belowdisplayskip{6pt}
\begin{align}
    L_{Relation}&=\sum_{(i,j) \in {\mathbb{K}^2}}{l(\frac{1}{\phi(t) }{\left \| t_{i}-t_{j}\right \|_{2}},\frac{1}{\phi(s) }{\left \| s^\prime_{i}-s^\prime_{j}\right \|_{2}})} \nonumber
    ,\\
    \phi(x)&=\frac{1}{\left | \mathbb{K}^2 \right | } \sum_{(i,j) \in {\mathbb{K}^2}}{{\left \| x_{i}-x_{j}\right \|_{2}}},
\end{align}
}

\par\noindent where $\mathbb{K}^2 = \left \{ \left (i,j \right ) | i\neq j,1\leqslant i,j\leqslant K \right  \}$, and $\phi(\cdot)$ is a normalization factor for distance, and $l$ denotes smooth L1 loss. 

\renewcommand\arraystretch{1.0}
\renewcommand\tabcolsep{5pt}
\begin{table*}[!ht]
\centering
\begin{tabular}{p{2.5cm}|p{1.6cm}<{\centering}p{1.4cm}<{\centering}|p{1.4cm}<{\centering}p{1.4cm}<{\centering}|p{1.4cm}<{\centering}p{1.4cm}<{\centering}}

\toprule
\multirow{2}*{\textbf{Method}} & \multicolumn{2}{c}{\textbf{Faster R-CNN Res101-50}} & \multicolumn{2}{|c|}{\textbf{RetinaNet Res101-50}} & \multicolumn{2}{c}{\textbf{FCOS Res101-50}} \\ \cline{2-7}
~ & \textbf{mAP} & \textbf{AP}$_{50}$ & \textbf{mAP} & \textbf{AP}$_{50}$ & \textbf{mAP} & \textbf{AP}$_{50}$ \\ 
\midrule
teacher & 56.3 & 82.8 & 57.3 & 81.9 & 58.4 & 81.6 \\ 
student & 54.2 & 82.2 & 55.4 & 80.9 & 56.1 & 80.2 \\ \midrule
Mimicking\cite{Li_2017_CVPR} & 55.5 & 82.3 & - & - & - & - \\ 
Fine-grained\cite{Wang_2019_CVPR} & 55.4 & 82.2 & 56.6 & 81.5 & - & - \\ 
Fitnet\cite{romero2014fitnets} & 55.1 & 82.2 & 55.8 & 81.4 & 57.0 & 80.3 \\ 
Ours & \textbf{56.5} & \textbf{82.6} & \textbf{57.9} & \textbf{82.0} & \textbf{58.4} & \textbf{81.3} \\ \bottomrule
\end{tabular}
\vspace{0.05cm}
\caption{Comparison with previous work on PASCAL VOC with different detection frameworks.  Some results are missing, as Mimicking and Fine-grained can only be applied to two-stage frameworks and anchor-based frameworks respectively.}
\label{VOC}
\end{table*}

\subsection{Response-based Distillation}
\cite{Yuan_2020_CVPR} proposes that the performance gain from the knowledge distillation mainly due to the regularization of the respond-based knowledge from the teacher model. However, performing the distillation on the whole output of the detection head is detrimental to the performance of the student model. We speculate that this may be caused by the imbalance of the positive and negative samples of the detection tasks and the noise introduced by too many negative samples. Recently, some detection distillation methods \cite{sun2020distilling,NIPS2017_6676} only distill the positive sample on the detection head, ignoring the regularization effect of the discriminative negative samples. Therefore, we designed distillation masks for the classification branch and regression branch based on selected GIs, which is proved more effective than only using GT labels as the distillation mask.

However, since the definition of outputs from the detector head varies from model to model, we propose a general framework to perform the distillation on the detection head for different model, as shown in Fig \ref{Response_distillation}. First of all, the distillation mask based on GIs is calculated as follows: 
{\setlength\abovedisplayskip{8pt}
\setlength\belowdisplayskip{-2pt}
\begin{align}
    M = F_{Assign}(GIs),
\end{align}
}

\par\noindent where function $F$ is label assignment algorithm, which is differ from model to model. It’s input is the GI boxes and it’s output is 1 when this output pixel is matched GI and 0 when it is not. \eg For RetinaNet, we use IoU between anchors and GIs to determine whether it is masked or not. For FCOS, all the outputs outside GIs are masked.

Then response-based loss can be expressed as follows:
{\setlength\abovedisplayskip{1pt}
\setlength\belowdisplayskip{5pt}
\begin{align}
    L_{Response} &= \frac{1}{N_{m}}\sum_{i=1}^{R} M_{i} \left (\alpha L_{cls}\left ( y_{t}^{i},y_{s}^{i}\right )+\beta L_{reg}\left ( r_{t}^{i},r_{s}^{i}\right )\right ), \nonumber
    \\
    N_{m} &= \sum_{i=1}^{R}{M_{i}},
\end{align}
}
\par\noindent in which $y_t$ $r_t$ are from teacher model while $y_s$ $r_s$ are from student model. $y_t$ $y_s$ are from output of the classification head. $r_t$ $r_s$ are from output of the regression head. $L_{cls}$ and $L_{reg}$ are the classification and regression loss function same as the task loss function of specific distilled model. It should be noted that, for two-stage detector, we distill the outputs of RPN instead for simplify.

\subsection{Overall loss function}
We trained the student model end-to-end, total loss for distilling student model is as follows:
{
\setlength\abovedisplayskip{8pt}
\setlength\belowdisplayskip{2pt}
\begin{align}
    L \!=\! L_{GT} \!+\! \lambda_{1}L_{Feature} \!+\! \lambda_{2}L_{Relation} \!+\! \lambda_{3}L_{Response},
\end{align}
}
\par\noindent where $L_{GT}$ is task loss for detection model, $\lambda_{1}$, $\lambda_{2}$, $\lambda_{3}$ are hyper-parameters to balance each loss in the same scale.
\section{Experiments}
In order to verify the effectiveness and robustness of our method, we conduct experiments on different detection frameworks, heterogeneous backbones and few classes detection with COCO and Pascal VOC dataset. Specifically, following the setting in \cite{NIPS2015_5638}, for the Pascal VOC dataset, we choose the 5k trainval images split in VOC 2007 and 16k trainval images split in VOC 2012 for training and 5k test images split in VOC 2007 for test. While for COCO, we choose the default 120k train images split for training and 5k val images split for test. All the distillation performances are evaluated in average precision (AP). 

We adopt the hyper-parameters $\{ K= 10, \lambda_{1}= 5\times 10^{-4}, \lambda_{2}= 40, \lambda_{3}= 1, \alpha = 0.1, \beta = 1\}$ for all experiments by diagnosing the initial loss of each knowledge type and ensuring that all losses are within the same scale. Unless specified, we use 2x learning schedule to train 24 epochs (180000 iterations) on COCO dataset and 17.4 epochs (18000 iterations) on VOC dataset for distillation. 

\renewcommand\arraystretch{1.0}
\renewcommand\tabcolsep{5.2pt}
\begin{table*}[!ht]
\centering
\begin{tabular}{l|c|c c|c c c|c|c c c}
\toprule
\textbf{Method} & \textbf{mAP} & \textbf{AP}$_{50}$ & \textbf{AP}$_{75}$ & \textbf{AP}$_{S}$ & \textbf{AP}$_M$ & \textbf{AP}$_L$ & \textbf{mAR} & \textbf{AR}$_S$ & \textbf{AR}$_M$ & \textbf{AR}$_L$ \\ 
\midrule
Retina-Res101 (teacher) & 38.1 & 58.3 & 40.9 & 21.2 & 42.3 & 51.1 & 54.4 & 34.1 & 59.1 & 70.5 \\ 
Retina-Res50 (student) & 36.2 & 55.8 & 38.8 & 20.7 & 39.5 & 48.7 & 52.1 & 33.7 & 55.3 & 68.6\\ 
+ Fitnet\cite{romero2014fitnets} & 37.4 & 57.1 & 40.0 & 20.8 & 40.8 & 50.9 & 53.5 & 33.6 & 57.4 & 69.4 \\
+ Fine-grained\cite{Wang_2019_CVPR} & 38.6 & 58.7 & 41.3 & 21.4 & 42.5 & 51.5 & 54.6 & 34.7 & 58.2 & 70.4 \\
+ Ours & 39.1 & 59.0 & 42.3 & 22.8 & 43.1 & 52.3 & 55.3 & 36.7 & 59.1 & 71.1 \\
Our gain & \textbf{+2.9} & \textbf{+3.2} & \textbf{+3.5} & \textbf{+2.1} & \textbf{+3.6} & \textbf{+3.6} & \textbf{+3.2} & \textbf{+3.0} & \textbf{+3.8} & \textbf{+2.5} \\ 
\midrule
FCOS-Res101 (teacher) & 41.0 & 60.3 & 44.2 & 24.6 & 44.8 & 52.8 & 58.8 & 39.8 & 63.9 & 74.0 \\ 
FCOS-Res50 (student) & 38.5 & 57.0 & 41.3 & 21.4 & 41.8 & 50.7 & 56.1 & 34.6 & 60.3 & 72.0 \\ 
+ Fitnet\cite{romero2014fitnets} & 39.9 & 58.6 & 43.1 & 23.1 & 43.4 & 52.2 & 57.3 & 36.6 & 61.5 & 73.1 \\
+ Ours & 42.0 & 60.4 & 45.5 & 25.6 & 45.8 & 54.2 & 59.9 & 39.9 & 64.3 & 75.8 \\
Our gain & \textbf{+3.5} & \textbf{+3.4} & \textbf{+4.2} & \textbf{+4.2} & \textbf{+4.0} & \textbf{+3.5} & \textbf{+3.8} & \textbf{+5.3} & \textbf{+4.0} & \textbf{+3.8} \\ 
\midrule
R-CNN-Res101 (teacher) & 39.6 & 60.6 & 43.1 & 22.7 & 43.3 & 51.9 & 53.3 & 32.8 & 57.5 & 67.7\\ 
R-CNN-Res50 (student) & 38.3 & 58.8 & 41.7 & 21.4 & 41.6 & 50.5 & 51.5 & 30.9 & 55.2 & 65.9\\ 
+ Fitnet\cite{romero2014fitnets}  & 38.9 & 59.5 & 42.4 & 21.9 & 42.2 & 51.6 & 52.3 & 32.5 & 55.4 & 66.7\\
+ Mimicking\cite{Li_2017_CVPR} & 39.6 & 60.1 & 43.3 & 22.5 & 42.8 & 52.2 & 52.9 & 33.1 & 56.1 & 67.6\\
+ Fine-grained\cite{Wang_2019_CVPR} & 39.3 & 59.8 & 42.9 & 22.5 & 42.3 & 52.2 & 52.4 & 32.2 & 55.7 & 67.9\\
+ Ours & 40.2 & 60.7 & 43.8 & 22.7 & 44.0 & 53.2 & 53.9 & 33.5 & 57.6 & 68.8\\
Our gain & \textbf{+1.9} & \textbf{+1.9} & \textbf{+2.1} & \textbf{+1.3} & \textbf{+2.4} & \textbf{+2.7} & \textbf{+2.4} & \textbf{+2.6} & \textbf{+2.4} & \textbf{+2.9} \\ 
\bottomrule
\end{tabular}
\vspace{0.05cm}
\caption{Results of the proposed GID on COCO dataset with different detection frameworks. }
\label{COCO_main_experiment}
\end{table*}

\subsection{Different detection frameworks}

We evaluate our method based on three state-of-the-art detection frameworks, anchor-based one-stage detector (RetinaNet), anchor-free one-stage detector (FCOS), and two-stage detector (Faster R-CNN). Among those three models, the distillation for feature-based and relation-based distillation is exactly the same. However, as the target definition for detection task in each model is different, the form of the response-based distillation loss is also different, \eg following the original loss setting, we choose IoU loss and smooth L1 loss for FCOS and RetinaNet separately.

As for the backbone, we choose shallower student backbone with similar architecture of teacher model. To be specific, we choose ResNet-50 based model as student model, and ResNet-101 based model as teacher model. As shown in Table \ref{VOC}, we compare our method with SOTA detection distillation methods on Pascal VOC. The results shows that our method outperforms the previous SOTA methods to a large extent and even surpasses the teacher model.

As shown in Table \ref{COCO_main_experiment}, we also perform experiments with COCO dataset. All student models get significant performance gains from the teacher by our method and reach a comparable result to the teacher models, \eg the ResNet-50 based RetinaNet student model gets 2.9 absolute gain in mAP, which totally recovers the performance drop due to the shallow backbone. Especially, our method achieves a further APs gain compared to other feature-based methods, since we treat each instance equally, regardless of the proportion of the instance in the feature map. Those results demonstrate that our method is applicable to a variety of widely used detection frameworks.

\subsection{Heterogeneous network backbones}

To further verify the generalization of our methods, instead of using homogeneous ResNet backbones for distillation, we introduce two heterogeneous backbones. Specifically, we take MobileNet-v2 \cite{DBLP:journals/corr/abs-1801-04381} based RetinaNet as student and ResNet-101 based one as teacher. As shown in Table \ref{h_backbone}, the lightweight MobileNet-V2 based detector gets a 2.5 absolute mAP gain even if the basic network module is different between student and teacher.

\begin{figure*}[!ht]
    \centering
    \subfigure[]{
        \includegraphics[width=1.5in, height=0.9in]{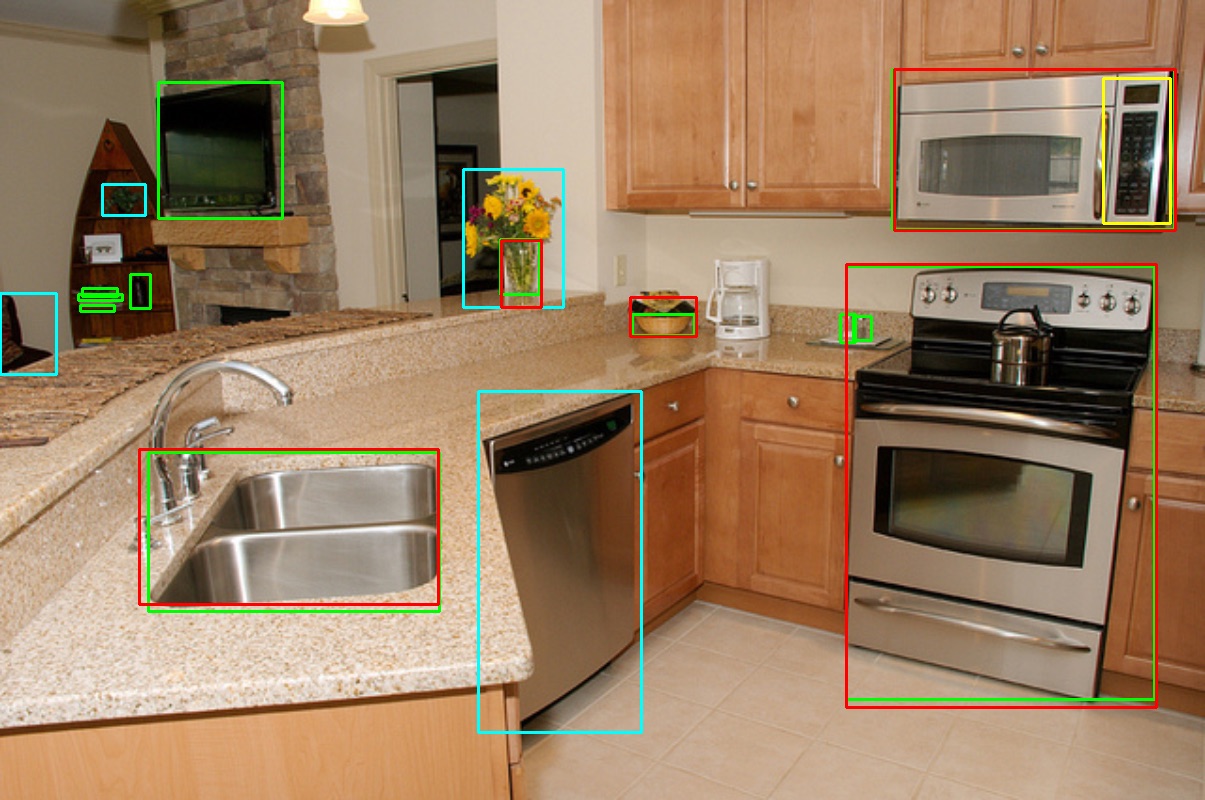}
        \label{4999_furniture}
    }
    \subfigure[]{
	    \includegraphics[width=1.5in, height=0.9in]{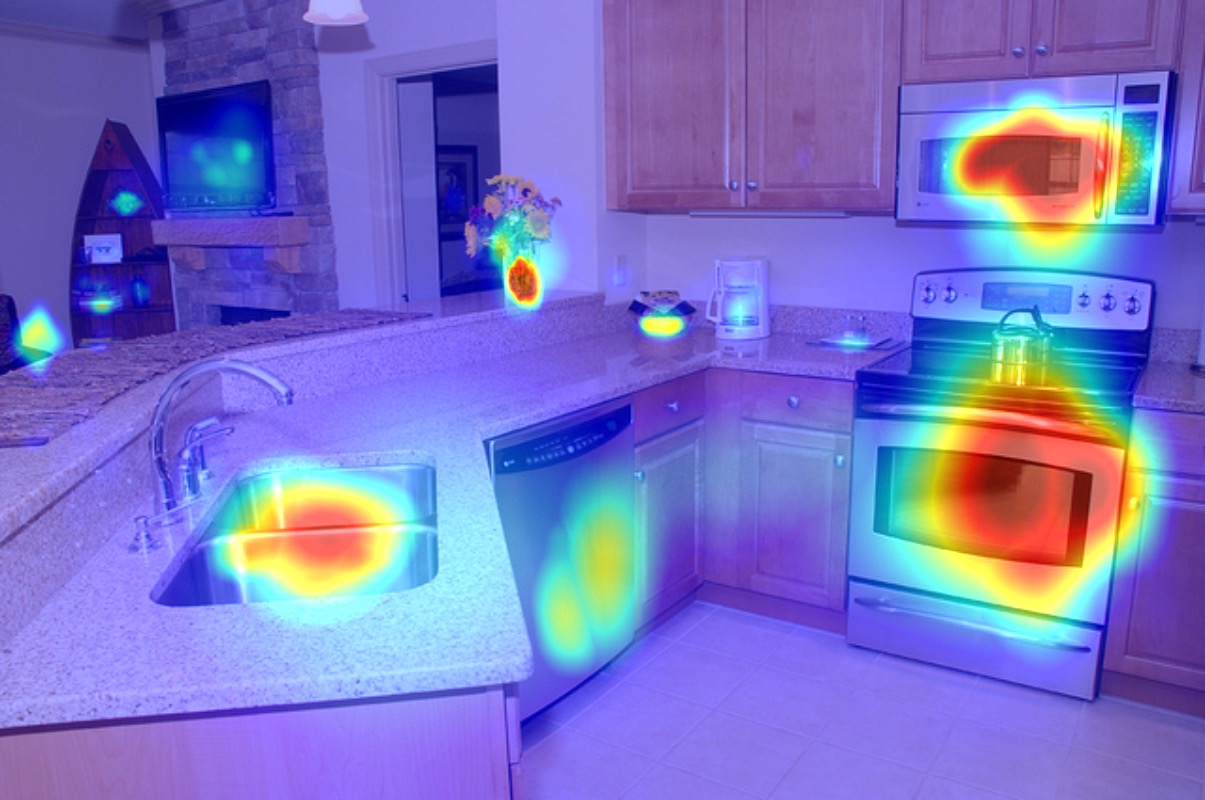}
	    \label{4999_furniture_heatmap}
    }
    \subfigure[]{
        \includegraphics[width=1.5in, height=0.9in]{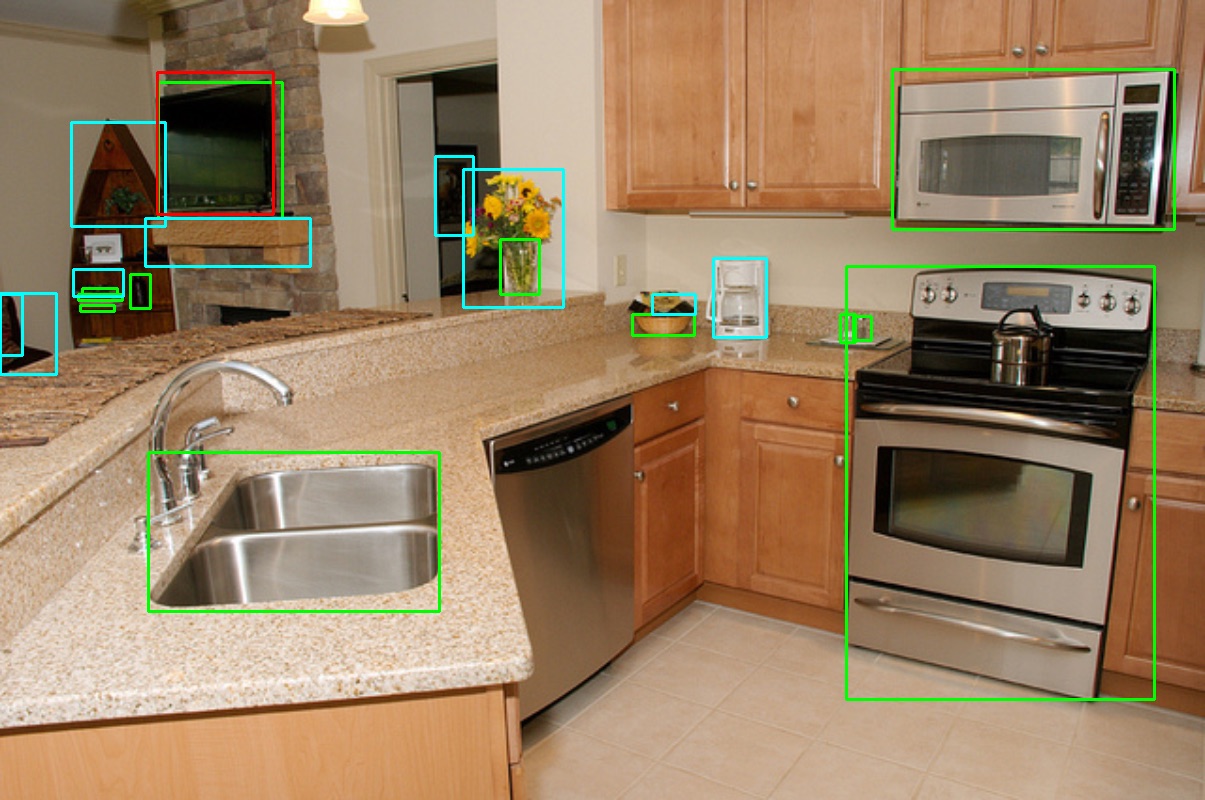}
        \label{90000_furniture}
    }
    \subfigure[]{
	    \includegraphics[width=1.5in, height=0.9in]{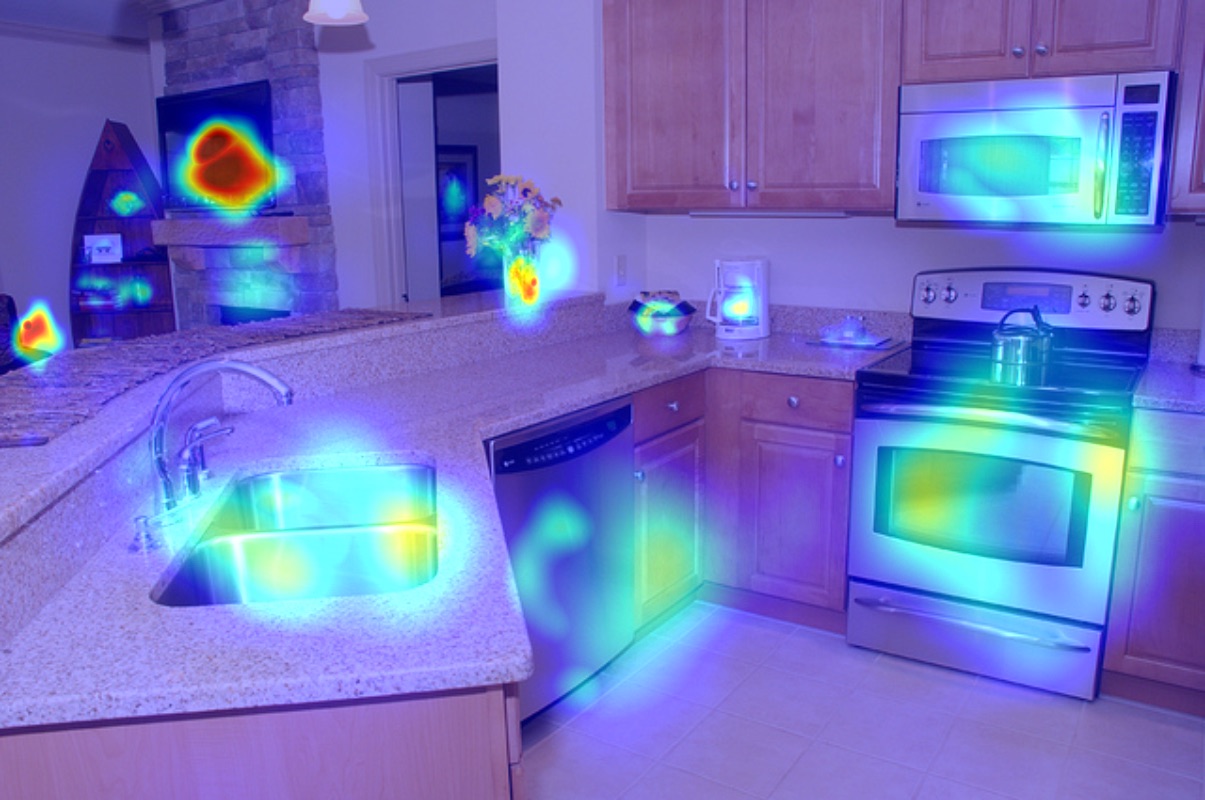}
	    \label{90000_furniture_heatmap}
    }
    \vspace{-0.3cm}
    \\    
    \subfigure[]{
    	\includegraphics[width=1.5in, height=0.9in]{visualization/new_color_4999_tennis.jpg}
    	\label{4999_tennis}
    }
    \subfigure[]{
	    \includegraphics[width=1.5in, height=0.9in]{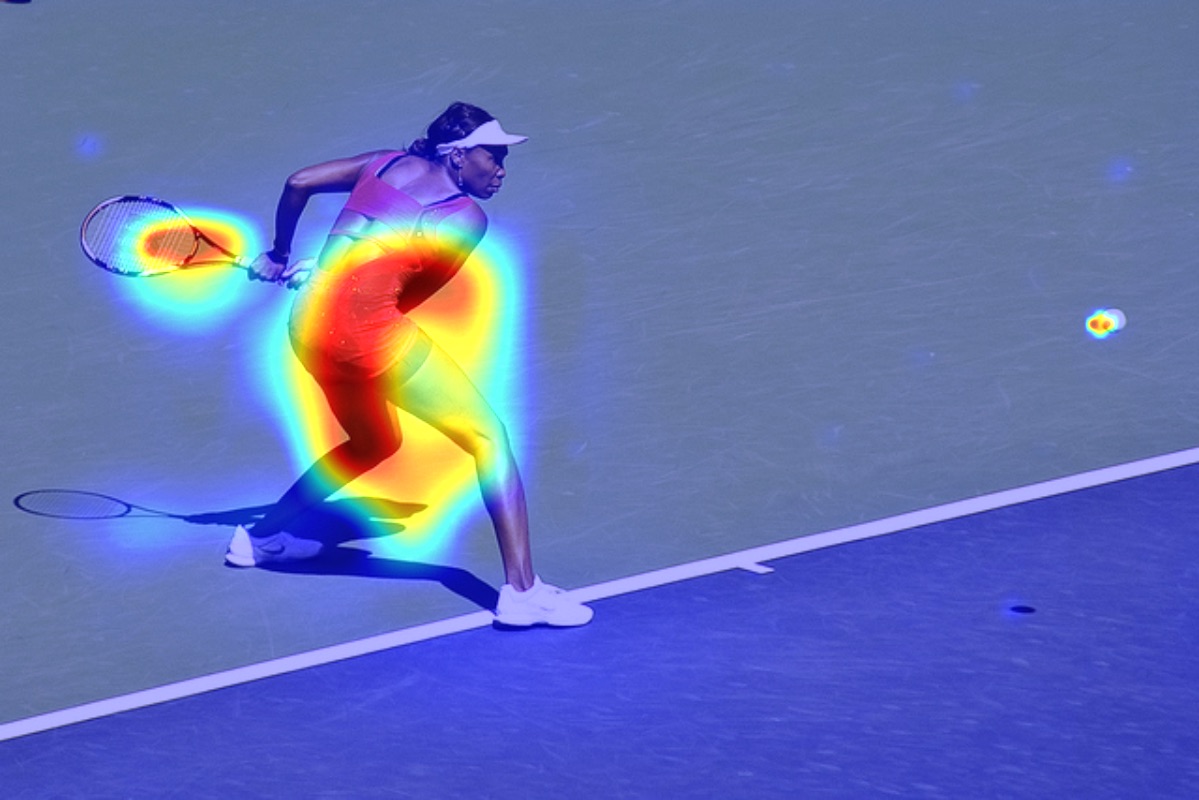}
	    \label{4999_tennis_heatmap}
    }
    \subfigure[]{
    	\includegraphics[width=1.5in, height=0.9in]{visualization/new_color_90000_tennis.jpg}
    	\label{90000_tennis}
    }
    \subfigure[]{
	    \includegraphics[width=1.5in, height=0.9in]{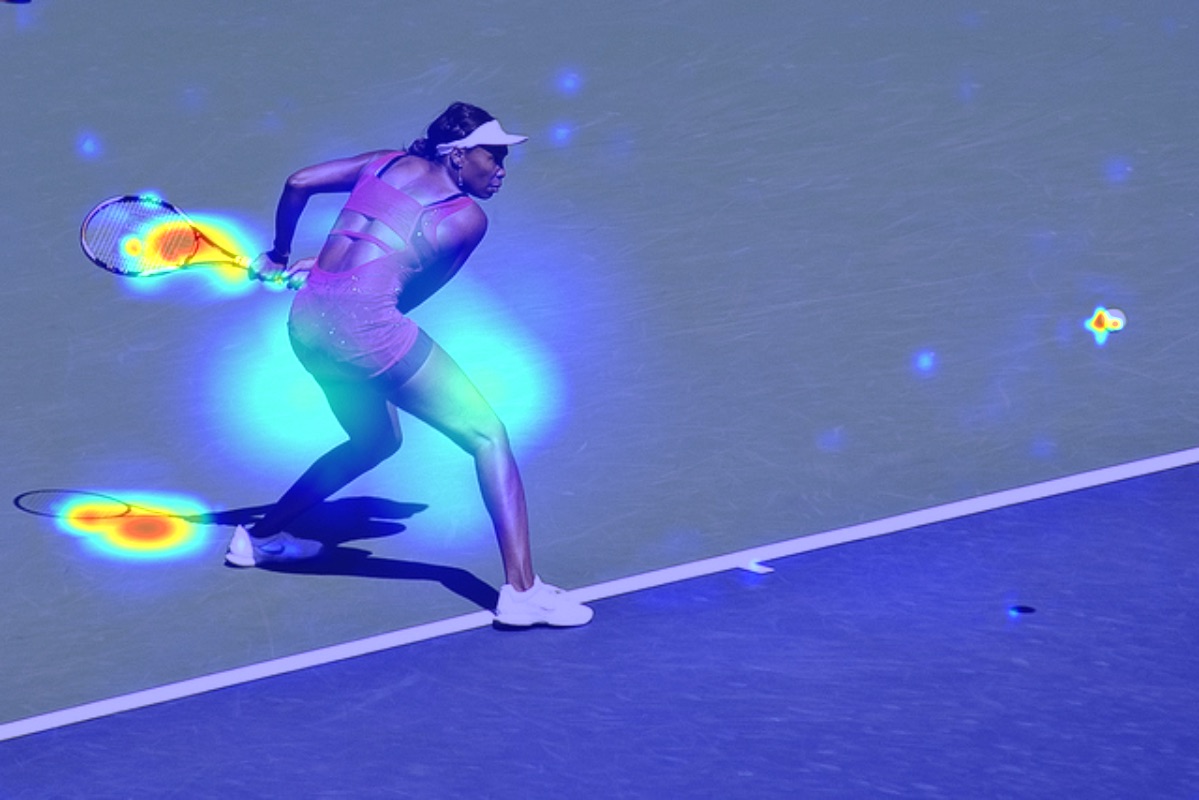}
	    \label{90000_tennis_heatmap}
    }
    \vspace{-0.1cm}
    \caption{Examples from COCO of GIs selected by GISM and the corresponding heat maps of the GI score distribution with RetinaNet-Res101-50 model. The instances from (a)(e) are calculated by the teacher and student with 5000 iterations, while the instances from (c)(g) are calculated by the teacher and student model with 90000 iterations. The \textcolor[RGB]{0,255,0}{green}, \textcolor[RGB]{255,0,0}{red}, \textcolor[RGB]{255,255,0}{yellow} and \textcolor[RGB]{0,255,255}{cyan} boxes denote ground truth, positive, semi-positive and negative instances respectively, as defined in Sec \ref{GI_gain}. For clarity, we only visualize the GI with top10 GI score.}
    \label{proposal_visualization}
\end{figure*}

\subsection{Distillation with fewer classes}

\cite{2020arXiv200203936M} points out that the information distilled is linear in the number of classes, so distillation is considerably less efficient for models with few classes. However, our method will adaptively select highly informative areas to distill and take advantage of all kinds of knowledge from the teacher model. Thus, we get impressive results, as shown in Table \ref{f_class}, when only a single class (person) is considered in the COCO dataset. The student model still exceeds the teacher model by a large margin in terms of few classes.

\renewcommand\arraystretch{1.2}
\renewcommand\tabcolsep{4.0pt}
\begin{table}[!ht]
\centering
\resizebox{8.3cm}{!}{
\begin{tabular}{l|c|c|c|c|c|c|c}
\toprule
\textbf{Model} & \textbf{Teacher} & \textbf{mAP} & \textbf{AP}$_{50}$ & \textbf{AP}$_{75}$ & \textbf{AP}$_{S}$ & \textbf{AP}$_{M}$ & \textbf{AP}$_{L}$ \\ 
\midrule
Retina-R101 & - & 38.1 & 58.3 & 40.9 & 21.2 & 42.3 & 51.1 \\ 
Retina-Mob & - & 31.0 & 48.9 & 32.7 & 16.4 & 33.8 & 42.6 \\ 
Retina-Mob & Retina-R101 & 33.5 & 51.9 & 35.5 & 19.2 & 36.9 & 44.3 \\
 & & \textbf{+2.5} & \textbf{+3.0} & \textbf{+2.8} & \textbf{+2.8} & \textbf{+3.1} & \textbf{+1.7} \\
\bottomrule
\end{tabular}}
\vspace{0.05cm}
\caption{GID results on COCO dataset with heterogeneous network backbones. }
\label{h_backbone}
\end{table}

\renewcommand\arraystretch{1.2}
\renewcommand\tabcolsep{4.0pt}
\begin{table}[!ht]
\centering
\resizebox{8.3cm}{!}{
\begin{tabular}{l|c|c|c|c|c|c|c}
\toprule
\textbf{Model} & \textbf{Teacher} & \textbf{mAP} &   
\textbf{AP}$_{50}$ & \textbf{AP}$_{75}$ & \textbf{AP}$_{S}$ & \textbf{AP}$_{M}$ & \textbf{AP}$_{L}$ \\
\midrule
Retina-R101 & - & 52.1 & 81.2 & 54.6 & 32.8 & 59.3 & 71.8 \\ 
Retina-R50 & - & 50.4 & 79.4 & 53.0 & 31.6 & 56.9 & 70.2 \\ 
Retina-R50 & Retina-R101  & 52.8 & 81.3 & 55.4 & 33.9 & 59.4 & 72.5 \\
& & \textbf{+2.4} & \textbf{+1.9} & \textbf{+2.4} & \textbf{+2.3} & \textbf{+2.5} & \textbf{+2.3} \\
\bottomrule
\end{tabular}}
\vspace{0.05cm}
\caption{GID results on COCO dataset with only Person class.}
\label{f_class}
\end{table}

\subsection{Analysis}
\subsubsection{Visualization of General Instances}

To better understand the general instances we selected to be distilled, we visualize the selected general instances and the corresponding heat maps of the GI score distribution from different training stages with RetinaNet-Res101-50 (ResNet-101 based teacher and ResNet-50 based student). As shown in Figure \ref{proposal_visualization}, the green boxes denote the ground-truth label, the other color boxes denote the general instance used to be distilled. The characteristic of general instances can be summarized into the following categories: 

\par\noindent\textbf{Key characteristic patches.} As shown in the yellow boxes in Fig \ref{4999_tennis} \ref{90000_tennis}, some key features of the athlete are selected such as the shoes and clothes, which are critical areas for distillation and are explained as visual concepts in \cite{Cheng_2020_CVPR}. 

\par\noindent\textbf{Extra instances.} Some confusing background areas with discriminative semantic information are also selected for distillation. For example, the oven-like machine and the shadow of the tennis racket shown in the cyan boxes in Fig \ref{4999_furniture} \ref{90000_tennis} are selected due to the inconsistent score distribution between the student model and teacher model, though some of those background instances do not include in the original 80 classes of the COCO dataset. 

\par\noindent\textbf{More informative positive instances.} Compared with ground truth boxes (green), we only choose a subset among those for distillation as shown in the red boxes in Fig \ref{proposal_visualization}. In contrast to Online Hard Example Mining (OHEM) \cite{DBLP:journals/corr/ShrivastavaGG16} which adaptively selects the hardest examples for training, in our method, the extremely hard instances are discarded for distillation, such as the small wine glasses in Fig \ref{4999_furniture}, which are hard to be detected from both the student and teacher model. While in the late stage of training, the majority of ground truth instances are simple for both teacher and student models and some of them are ignored for distillation. As shown in Fig \ref{90000_furniture} \ref{90000_tennis}, the athlete and the microwave oven are neglected, only some positive instances which are hard to learn from teacher remain to be distilled, like the tennis ball. 

\noindent\textbf{Time-varying distillation tendency.} With the training goes on, the focus of the distillation targets has shifted from simple positive instances to small confusing patches, as shown in the heat maps of GI score in Fig \ref{4999_furniture_heatmap} \ref{4999_tennis_heatmap} from iteration 5000 and Fig \ref{90000_furniture_heatmap} \ref{90000_tennis_heatmap} from iteration 90000. This distillation tendency is similar to a human cognitive process. The above scenarios demonstrate that the GISM will adaptively choose the most informative and discriminative patches for distillation during training. 

\subsubsection{Performance gain from General Instance} \label{GI_gain}

To further analyze the contribution of each type of general instances and verify the effectiveness of GISM, we perform experiments on each type of general instances. We introduce an index named intersection over proposals (IoP) to help us separate those GIs:
{
\setlength\abovedisplayskip{9pt}
\setlength\belowdisplayskip{3pt}
\begin{align}
    IoP=\frac{area(GI \cap GT)}{area(GI)}
\end{align}
}

Then we define each type of GIs as follows:
{\setlength\abovedisplayskip{6pt}
\setlength\belowdisplayskip{3pt}
\begin{align}
GI=\left\{
\begin{array}{lll}
\text{Pos} & \text{IoU}>0.5 \\ 
\text{Semi-Pos} & \text{IoU} \leq 0.5,\ \text{IoP} > 0.7 \\ 
\text{Neg} & \text{IoU} \leq 0.5,\ \text{IoP} < 0.3 \\
\end{array}\right. ,
\end{align}
}
\par\noindent where the IoU means the intersection over union between GI and GT. Pos, Semi-Pos, Neg are short for positive, semi-positive and negative instances respectively. Besides, in ablation study, we ignore those instances with an IoP between 0.3 to 0.7 to analyse the contribution of each part of the general instances more clearly.

\renewcommand\arraystretch{1.2}
\renewcommand\tabcolsep{4.8pt}
\begin{table}[!ht]
\centering
\resizebox{8.3cm}{!}{
\begin{tabular}{l|c|c|c|c|c|c}
\toprule
\multirow{2}*{Model} & \multicolumn{6}{c}{RetinaNet Res101-50} \\\cline{2-7}
  & GT & \multicolumn{5}{c}{GI} \\
\hline
Positive & - & $\surd$ & - & - & $\surd$ & $\surd$  \\ 
Semi-Positive & - & - & $\surd$ & - & $\surd$ & $\surd$  \\ 
Negative & - & - & - & $\surd$ & - & $\surd$  \\
\hline
mAP & 38.5 & 38.8 & 38.6 & 38.2 & 39.0 & \textbf{39.1} \\
\bottomrule
\end{tabular}}
\vspace{0.001cm}
\caption{Ablation Study for each type of General Instances, including positive, semi-positive and negative Instance. GT denotes that we use GT instance for distillation. }
\label{ablation_GI}
\end{table}

As shown in Table \ref{ablation_GI}, distilling each type of instance can all bring performance gain to the student model, while combining all three types can achieve the best performance. One thing that's very noticeable is that performing distillation only on negative instances is still beneficial to the student model, which is a strong evidence that our approach effectively selects the useful information from the background area while filters detrimental knowledge. Moreover, even if positive instances chosen by GISM is only a subset of GT instances, the result from positive instances surpasses that from GT instance, which indicate that some extreme hard or simple GT instances for both teacher and student will be detrimental for distillation. Besides, it is still effective to use the GT region for distillation. The essence is that the GT region is still the most informative and discriminative in the early stage of training. However, ignoring the hidden information in the background will make the student model fail to achieve better performance.
\begin{figure}[!h]
  \centering
  \vspace{-0.38cm}
  \includegraphics[width = \linewidth]{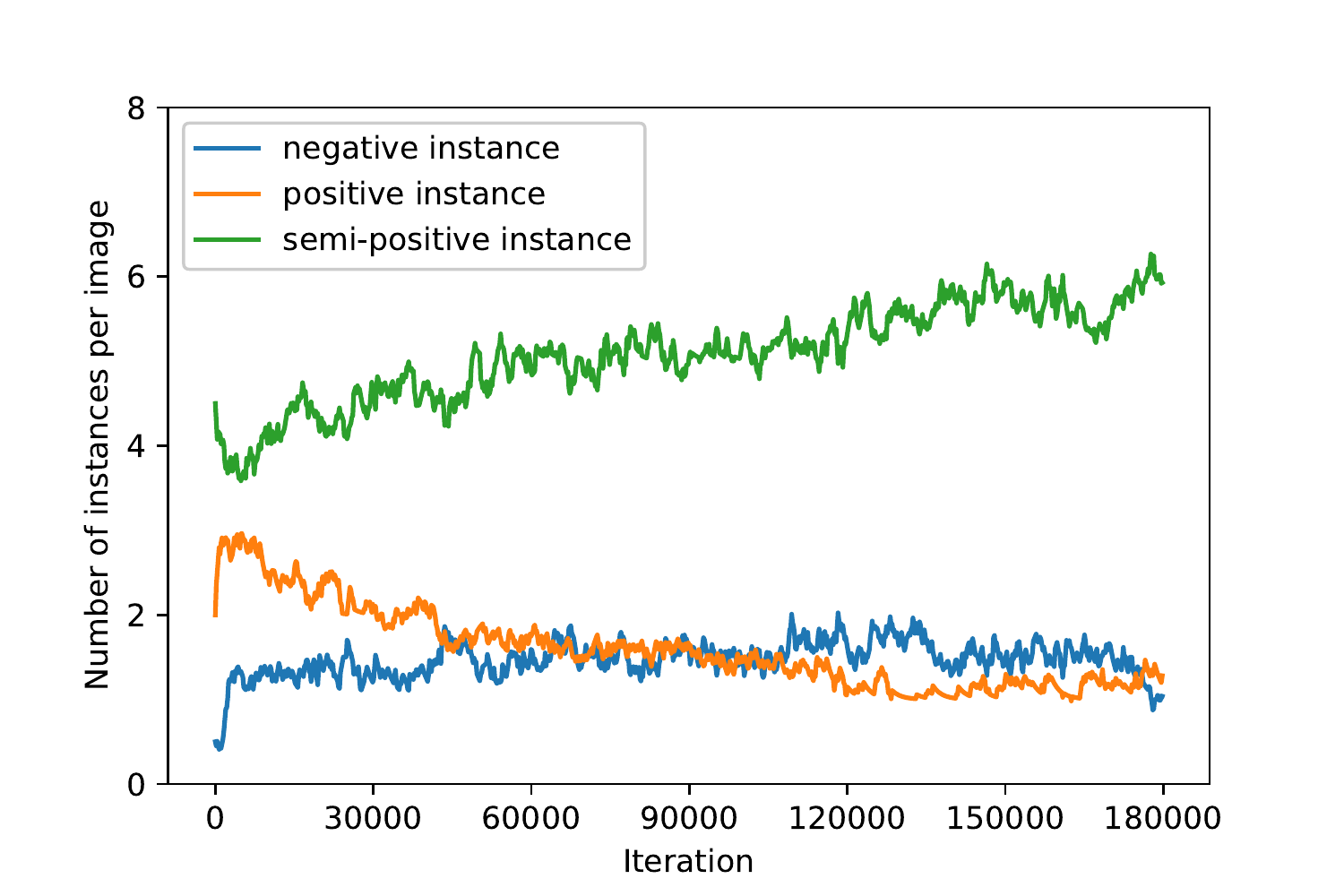} 
  \caption{The curve of the instances number with training iteration on RetinaNet-Res101-50. For a better demonstration, we perform exponential smoothing on the instance number with $\alpha = 0.9$.}
  \label{instance_fig}
\end{figure}

As shown in Fig \ref{instance_fig}, We drew the curves of these three types instances with the number of iterations. It can be seen that the proportion of positive distillation instances in the late training period is reduced, indicating that the improvement in the late training period is mainly brought by the discriminative background area and key characteristic patches.

\parskip=-2pt
\subsubsection{Performance gain from various knowledge}

In this subsection, we conduct several ablation experiments to understand how each type of knowledge component contributes to the final performance. As shown in Table \ref{ablation_distill_part}, each distillation component improves the performance, especially for the feature-based and response-based knowledge which brings improvement about 1.7 mAP gains. The combination of the above knowledge achieves the best results, which brings about another 1.2 mAP gains compared with the best single component.
That is to say, although the three types of knowledge contain some overlapping parts, we take advantage of the unique parts among them.
\renewcommand\arraystretch{1.2}
\renewcommand\tabcolsep{4.8pt}
\begin{table}[!ht]
\centering
\resizebox{8.3cm}{!}{
\begin{tabular}{l|c|c|c|c|c|c}
\toprule
Model & Res50 & \multicolumn{5}{c}{RetinaNet Res101-50} \\ 
\hline
Feature-based & - &  $\surd$ & - & - & $\surd$ & $\surd$ \\ 
Relation-based & - &  - & $\surd$ & - & - & $\surd$ \\ 
Response-based & - &  - & - & $\surd$ & $\surd$ & $\surd$ \\ 
\hline
mAP & 36.2 & 37.9 & 37.3 & 37.9 & 38.7 & \textbf{39.1} \\
\bottomrule
\end{tabular}}
\vspace{0.001cm}
\caption{Ablation Study for each part of the distillation loss.}
\label{ablation_distill_part}
\end{table}

\parskip=-2pt
\subsubsection{Varying top $K$ for GISM}

We investigate the influence of different GISM hyperparameter top $K$ with RetinaNet-Res101-50 model. As shown in Table \ref{ablation_topk}, when $K = 0$, no general instance is chosen, thus distillation loss is not applied. As the $K$ increases, the student model gets a significant mAP gain, even with 5 distilled GIs, which is strong evidence that our approach selects the most worthy instances for distillation. The performance becomes stable and optimal when the $K$ is at the range of 10 to 100, while it starts decreasing when $K$ still goes further, which is mainly because the informative and discriminative GIs are overwhelmed by trivial instances.
\renewcommand\arraystretch{1.2}
\renewcommand\tabcolsep{5.0pt}
\begin{table}[!ht]
\centering
\begin{tabular}{c|cccccc}
\toprule
Top $K$ & 0 & 5 & 10 & 40 & 100 & 300 \\
\midrule
mAP & 36.2 & 38.9 & \textbf{39.1} & 39.0 & \textbf{39.1} & 38.6 \\
\bottomrule
\end{tabular}
\vspace{0.08cm}
\caption{Hyperparameter analysis of top $K$ GI score.}
\label{ablation_topk}
\end{table}

\section{Conclusion}
We propose GID framework that adaptively selects the most discriminative instances between teacher and student for distillation. Besides, our method effectively improves the performance of modern detection frameworks with feature-based, relation-based and response-based knowledge, and is applicable to various detection frameworks. The ablation study demonstrates that imitating some of the GT instances will do harm to the performance while even some instances from backgrounds can be helpful, which will give some insights for future distillation works.

\vspace{1.0em}
\par\noindent\textbf{Acknowledgment.} This paper is supported by the National Key R\&D Plan of the Ministry of Science and Technology (“Grid function expansion technology and equipment for community risk prevention”, Project No. 2018YFC0809704).

{\small
\bibliographystyle{ieee_fullname}
\bibliography{egpaper_final}
}

\end{document}


\title{General Instance Distillation for Object Detection (Supplementary Material)}

\author{First Author\\
Institution1\\
Institution1 address\\
{\tt\small firstauthor@i1.org}
\and
Second Author\\
Institution2\\
First line of institution2 address\\
{\tt\small secondauthor@i2.org}
}

\maketitle

\section{Implementation Details}

We use a private codebase based on \textit{pytorch} designed for detection tasks to conduct our experiments. For more specific implementation details, you can go through the attached log files. Some summarized configurations are listed below.
\subsection{Overall Configuration}

All experiments are performed on NVIDIA GeForce RTX 2080Ti 8 GPUs with parallel acceleration. We adopt Stochastic Gradient Descend (SGD) as an optimization method with the weight decay of 0.0001 and the momentum of 0.9. Besides, we set batch size to 16, allocating 2 images per GPU. We load the ImageNet pre-trained ResNet-50 and ResNet-101 model as the initial weight for all the detector backbone in our experiments.

As for COCO dataset, followed the common practice, all models are trained for 2x for 24 epochs, i.e., 180k iterations with an initial learning rate of 0.01, which is then divided by 10 at 120k and 160k iterations. As for Pascal VOC dataset, all models are trained for 17.4 epochs at a total of 18k iterations, and the learning rate is divided by 10 at 12k and 16k iterations.

In addition, horizontal image flipping is utilized in data augmentation for both COCO and VOC datasets. For COCO dataset, single-scale training is performed, and the shortest edge is resized to 800 pixels. For VOC dataset, multi-scale training is applied, and the shortest edge is resized to (480, 512, 544, 576, 608, 640, 672, 704, 736, 768, 800) randomly. In the testing stage, the shortest edge of the test image is resized to 800 pixels for both COCO and VOC dataset.

Moreover, for VOC dataset, besides the conventional $\textbf{AP}_{50}$ metric that considers the threshold of IoU $\geq 0.5$, we adopt the overall $\textbf{mAP}$ metric, same as the definition in COCO dataset, averaging over the 10 IoU thresholds from 0.5 to 0.95.

\subsection{Model Configuration}

\textbf{For RetinaNet,} we adopt the hyper-parameter setting same as the optimal setting in original paper \cite{Lin_2017_ICCV}. We apply Focal Loss with $\alpha = 0.25$ and $\gamma = 2.0 $ for classification and Smooth L1 Loss with $\beta = 0.1$ for box regression. 

\textbf{For FCOS,}  we apply Focal Loss with $\alpha = 0.25$ and $\gamma = 2.0 $ for classification and IOU Loss for box regression and Binary Cross Entropy Loss for centerness branch. 

\textbf{For Faster-RCNN,} we apply FPN based version same as the one in \cite{Lin_2017_CVPR}. 

\subsection{Distillation Configuration}
     The hyper-parameters have already mentioned in our paper, i.e. $\{ K= 10, \lambda_{1}= 5\times 10^{-4}, \lambda_{2}= 40, \lambda_{3}= 1, \alpha = 0.1, \beta = 1\}$. Besides, we adopt 3x3 convolutional layers as the linear adaptation function in formula (5). The pooler resolution for ROIAlign algorithm used to extract the GI feature is set as (10, 10) and sampling ratio is set as 2.
     
\section{Additional Visualization Results}

\begin{figure*}[!t]
\subfigure{
\begin{minipage}[t]{0.01\textwidth}
\vspace{-70pt}
\rotatebox[]{90}{\textbf{Baseline}}

\vspace{75pt}
\\
\rotatebox[]{90}{\textbf{Distilled}}
\vspace{5pt}
\end{minipage}
}
\setcounter{subfigure}{0}
\subfigure[]{
\begin{minipage}[t]{0.23\textwidth}
\centering
\includegraphics[width=1\textwidth, height=1.5in]{visualization_supplementary_material/000000064495.jpg}\\
\includegraphics[width=1\textwidth, height=1.5in]{visualization_supplementary_material/distill_000000064495.jpg}
\end{minipage}%
\label{parking}
}%
\subfigure[]{
\begin{minipage}[t]{0.23\textwidth}
\centering
\includegraphics[width=1\textwidth, height=1.5in]{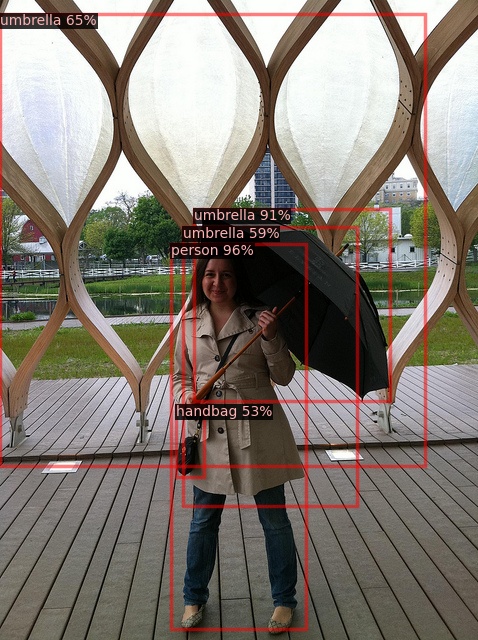}
\\
\includegraphics[width=1\textwidth, height=1.5in]{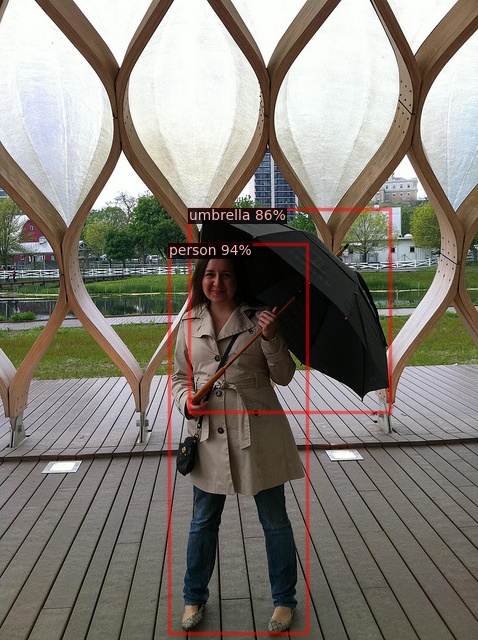}
\end{minipage}%
\label{umbrella}
}%
\subfigure[]{
\begin{minipage}[t]{0.23\textwidth}
\centering
\includegraphics[width=1\textwidth, height=1.5in]{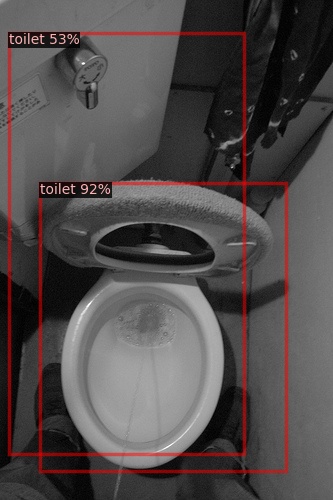}
\\
\includegraphics[width=1\textwidth, height=1.5in]{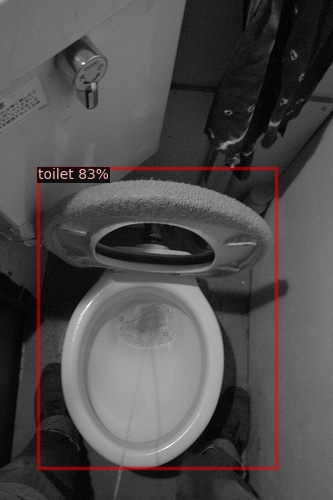}
\end{minipage}
\label{toilet}
}%
\subfigure[]{
\begin{minipage}[t]{0.23\textwidth}
\centering
\includegraphics[width=1\textwidth, height=1.5in]{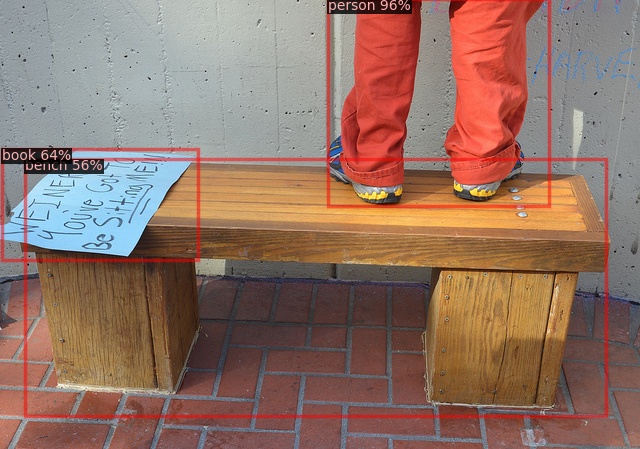}
\\
\includegraphics[width=1\textwidth, height=1.5in]{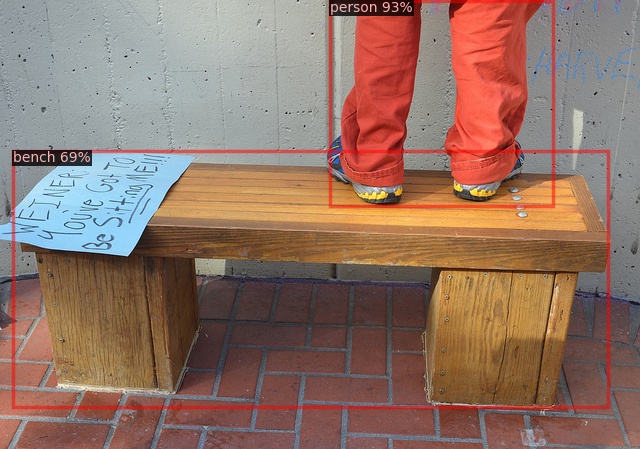}
\end{minipage}
\label{bench}
}


\subfigure{
\begin{minipage}[t]{0.01\textwidth}
\vspace{-70pt}
\rotatebox[]{90}{\textbf{Baseline}}

\vspace{75pt}
\\
\rotatebox[]{90}{\textbf{Distilled}}
\vspace{5pt}
\end{minipage}
}
\setcounter{subfigure}{4}
\subfigure[]{
\begin{minipage}[t]{0.23\textwidth}
\centering
\includegraphics[width=1\textwidth, height=1.5in]{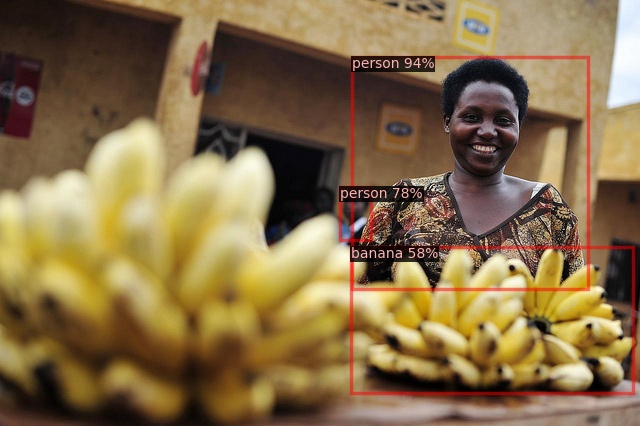}\\
\includegraphics[width=1\textwidth, height=1.5in]{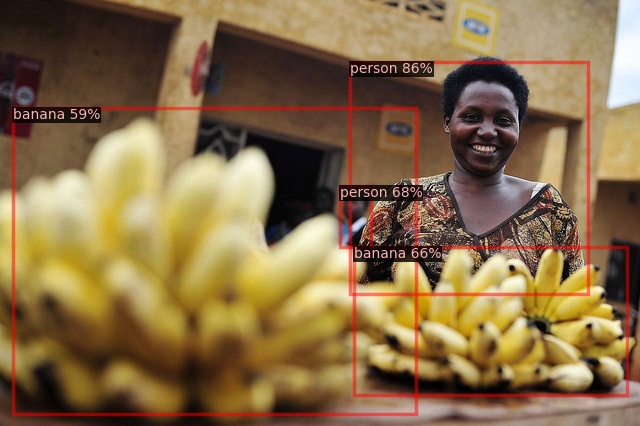}
\end{minipage}%
\label{banana}
}%
\subfigure[]{
\begin{minipage}[t]{0.23\textwidth}
\centering
\includegraphics[width=1\textwidth, height=1.5in]{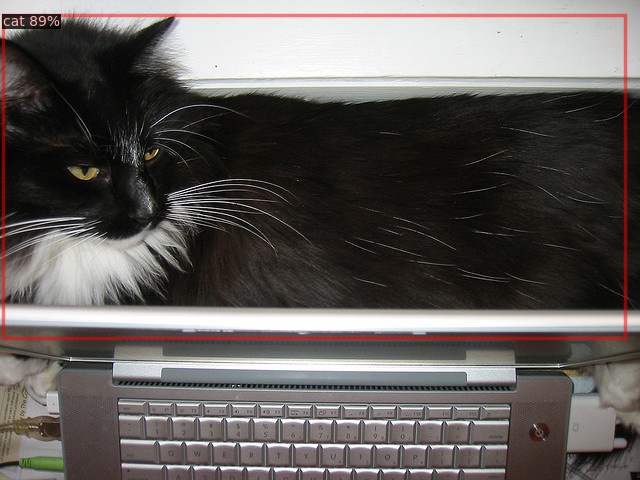}
\\
\includegraphics[width=1\textwidth, height=1.5in]{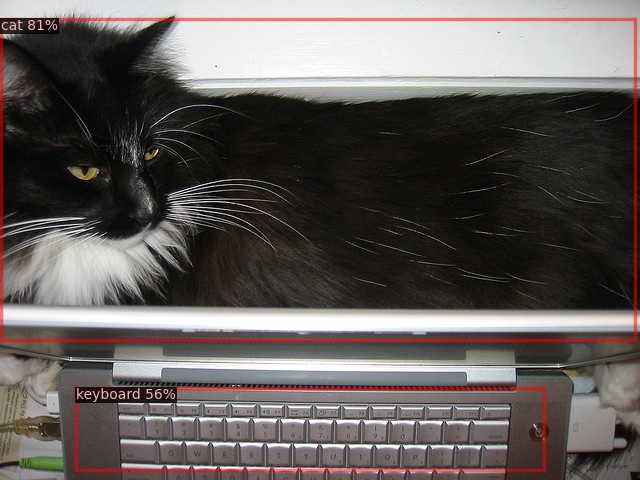}
\end{minipage}%
\label{cat}
}%
\subfigure[]{
\begin{minipage}[t]{0.23\textwidth}
\centering
\includegraphics[width=1\textwidth, height=1.5in]{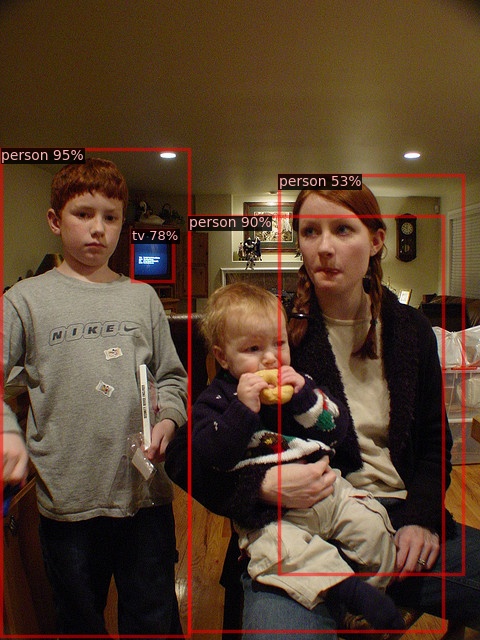}
\\
\includegraphics[width=1\textwidth, height=1.5in]{visualization_supplementary_material/distill_000000084674.jpg}
\end{minipage}
\label{person}
}%
\subfigure[]{
\begin{minipage}[t]{0.23\textwidth}
\centering
\includegraphics[width=1\textwidth, height=1.5in]{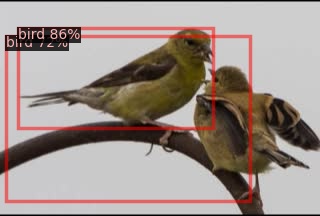}
\\
\includegraphics[width=1\textwidth, height=1.5in]{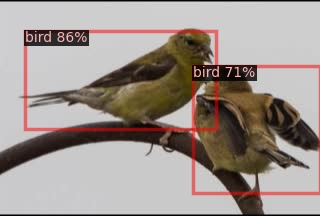}
\end{minipage}
\label{bird}
}

\caption{Qualitative analysis of distillation benefits. We visualize the predicted result on COCO dataset. The confidence threshold of bounding box visualization is set as 0.5. The first and third rows are the results of the baseline model, and the second and fourth rows are the results of the distilled model.}
\label{result_visualization}
\end{figure*}

To qualify the performance gain from the proposed GID, we visualize the predicted boxes from both the baseline model (RetinaNet-Res50 with 36.2 mAP) and distilled model (RetinaNet-Res101-50 with 39.1 mAP). As shown in Fig \ref{result_visualization}, we can directly feel the improvement brought by distillation. The improvement of the student model with teacher supervision can be summarized into the following aspects: \textbf{Less false positive samples.} As shown in Fig \ref{parking} \ref{toilet} \ref{umbrella}, the baseline model repeatedly detect the parking meters, toilets and umbrellas which are unfortunately not filtered by NMS, while the distilled model clearly detect those targets, maybe due to the successful transferring of the relation-based knowledge among instances. Besides, in Fig \ref{bench}, the paper is wrongly detected as a book by the baseline model. \textbf{Less missing samples.} As shown in Fig \ref{banana} \ref{cat} The distilled model correctly detect the banana in the front and the keyboard, while the baseline model ignores those targets. \textbf{More accurate localization.} As shown in Fig \ref{person} \ref{bird}, the distilled model predicts rather accurate bounding boxes for the person and bird than the baseline model, which demonstrate that the student gains a better localization ability.

{\small
\bibliographystyle{ieee_fullname}
\bibliography{egpaper_final}
}